\documentclass[letterpaper, 10 pt, conference]{ieeeconf}
\IEEEoverridecommandlockouts

\usepackage[hidelinks]{hyperref}
\usepackage{subcaption}
\usepackage{booktabs}       
\usepackage{placeins}
\usepackage[nocompress]{cite}
\usepackage{graphicx}
\usepackage[acronym,shortcuts]{glossaries}
\usepackage{comment}
\usepackage{enumerate}
\usepackage[capitalise]{cleveref}
\usepackage[table,xcdraw,dvipsnames]{xcolor}
\usepackage{siunitx}
\usepackage{bm} 
\usepackage{amsmath,amssymb,amsfonts}
\usepackage[english]{babel}
\usepackage{multirow}
\usepackage{pifont}
\usepackage{algorithm}
\usepackage{algpseudocode}
\usepackage{cuted}
\usepackage{cases}
\usepackage{textcomp}
\usepackage{xcolor}
\usepackage{fancyhdr}
\usepackage{tikz}
\usepackage{eso-pic}
\def\BibTeX{{\rm B\kern-.05em{\sc i\kern-.025em b}\kern-.08em
    T\kern-.1667em\lower.7ex\hbox{E}\kern-.125emX}}
\usepackage{eso-pic}
\newcommand\AtPageUpperCenterNotice[1]{%
  \AtPageUpperLeft{
    \put(\LenToUnit{0.5\paperwidth},\LenToUnit{-3cm}){\makebox[0pt]{#1}}
  }
}
\AddToShipoutPictureBG*{%
  \AtPageUpperCenterNotice{%
    \parbox[b][2cm][c]{\paperwidth}{
      \centering
      \fontsize{12}{14}\selectfont
      \color{gray!50}
      This paper has been accepted for publication in \\
International Conference on Intelligent Robots and Systems (IROS), Hangzhou 2025. \copyright{}IEEE.
    }
  }
}

\AddToShipoutPictureBG*{
  \AtPageLowerLeft{%
    \raisebox{25pt}{\makebox[\paperwidth]{\begin{minipage}{21cm}\centering
    \fontsize{10}{8}\selectfont
    \textcolor{gray!50}{%
      \copyright{} 2025 IEEE. Personal use of this material is permitted.
      Permission from IEEE must be obtained for all other uses, in any current or future media,
      including reprinting/republishing this material for advertising or promotional purposes,
      creating new collective works, for resale or redistribution to servers or lists,
      or reuse of any copyrighted component of this work in other works.
    }
    \end{minipage}}}%
  }
}

\definecolor{lightgray}{gray}{0.9}

\def\BibTeX{{\rm B\kern-.05em{\sc i\kern-.025em b}\kern-.08em
    T\kern-.1667em\lower.7ex\hbox{E}\kern-.125emX}}
\title{\LARGE \bf \vspace{6mm}
FSDP: Fast and Safe Data-Driven Overtaking Trajectory Planning for Head-to-Head Autonomous Racing Competitions
}

\author{Cheng Hu$^{1*}$, Jihao Huang$^{1*}$, Wule Mao$^{1*}$, Yonghao Fu$^{1*}$, Xuemin Chi$^{1}$, Haotong Qin$^{2\dagger}$, \\ Nicolas Baumann$^{2}$, Zhitao Liu$^{1}$, Michele Magno$^{2}$, and Lei Xie$^{1\dagger}$
\thanks{$^*$ \textbf{Equal contribution}.} 
\thanks{$1$ Authors are associated with the Department of Control Science and Engineering, Zhejiang University.}
\thanks{$2$ Authors are associated with the Center for Project-Based Learning, D-ITET, ETH Zurich.}
\thanks{$^\dagger$ Corresponding Authors {\tt\small {haotong.qin@pbl.ee.ethz.ch, lxie@iipc.zju.edu.cn}}.}}

\begin{document}

\newacronym{lidar}{LiDAR}{Light Detection and Ranging}
\newacronym{radar}{RADAR}{Radio Detection and Ranging}
\newacronym{mpc}{MPC}{Model Predictive Control}
\newacronym{mpcc}{MPCC}{Model Predictive Contouring Control}
\newacronym{iot}{IoT}{Internet of Things}
\newacronym{bev}{BEV}{Bird's-Eye View}

\newacronym{gpu}{GPU}{Graphics Processing Unit}
\newacronym{fps}{FPS}{Frames Per Second}
\newacronym{kf}{KF}{Kalman Filter}
\newacronym{ads}{ADS}{Autonomous Driving Systems}
\newacronym{iac}{IAC}{Indy Autonomous Challenge}
\newacronym{fsd}{FSD}{Formula Student Driverless}
\newacronym{rrt*}{RRT*}{Rapidly exploring Random Tree Star}
\newacronym{mgbt}{MGBT}{Multilayer Graph-Based Trajectory}
\newacronym{ftg}{FTG}{Follow The Gap}
\newacronym{gbo}{GBO}{Graph-Based Overtake}
\newacronym{rbf}{RBF}{Radial Basis Function}
\newacronym{map}{MAP}{Model- and Acceleration-based Pursuit}
\newacronym{sqp}{SQP}{Sequential Quadratic Programming}
\newacronym{roc}{RoC}{Region of Collision}
\newacronym{cpu}{CPU}{Central Processing Unit}
\newacronym{cots}{CotS}{Commercial off-the-Shelf}
\newacronym{obc}{OBC}{On-Board Computer}

\newacronym{sota}{SOTA}{State-of-the-Art}
\newacronym{qp}{QP}{Quadratic Programming}
\newacronym{A2RL}{A2RL}{Abu Dhabi Racing League}
\newacronym{GP}{GP}{Gaussian Process}
\newacronym{FITC}{FITC}{Fully Independent Training Conditional}
\newacronym{VFE}{VFE}{Variational Free Energy}
\newacronym{LQR}{LQR}{Linear Quadratic Regulator}
\newacronym{FSDP}{FSDP}{\textit{Fast and Safe Data-Driven Planner}}
\newacronym{gp}{GP}{Gaussian Process}
\newacronym{sgp}{SGP}{Sparse Gaussian Process}
\newacronym{VESC}{VESC}{Vedder Electronic Speed Controller}

\def\namealgo{RLPP}
\def\sim2real{Sim-to-Real}

\maketitle
\thispagestyle{fancy}

\begin{strip}
\vspace{-2.5cm}
\centering
\includegraphics[width=\textwidth]{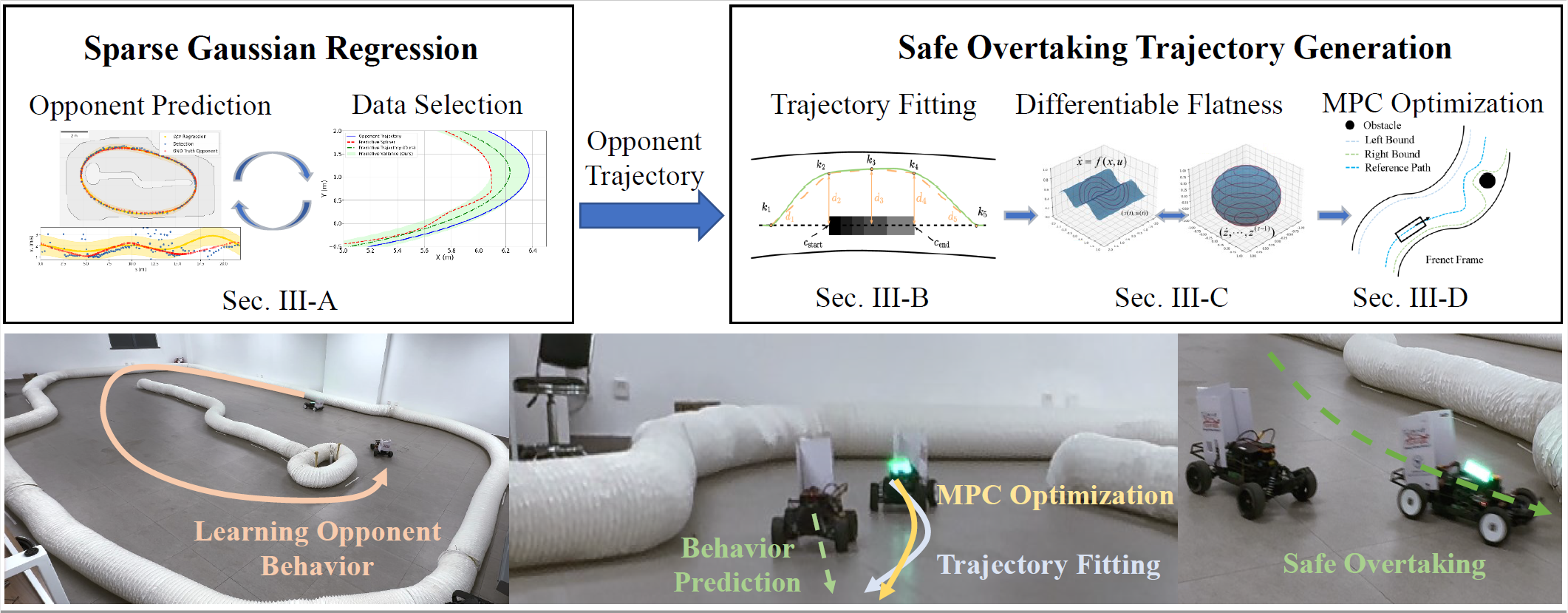}
\captionof{figure}{Framework of the proposed \emph{Fast and Safe Data-Driven Planner}:
The proposed method first employs the \gls{sgp} to learn the opponent's behavior and predict its trajectory. 
Based on this prediction, a bi-level \gls{qp} planner generates an overtaking trajectory that ensures both kinematic feasibility and safety. 
}
\label{fig:graphical_abstract}
\vspace{-0.475cm}
\end{strip}

\pagestyle{empty}

\setlength{\tabcolsep}{3pt}
\glsresetall
\begin{abstract}
Generating overtaking trajectories in autonomous racing is a challenging task, as the trajectory must satisfy the vehicle's dynamics and ensure safety and real-time performance running on resource-constrained hardware.
This work proposes the Fast and Safe Data-Driven Planner to address this challenge. Sparse Gaussian predictions are introduced to improve both the computational efficiency and accuracy of opponent predictions. 
Furthermore, the proposed approach employs a bi-level quadratic programming framework to generate an overtaking trajectory leveraging the opponent predictions. 
The first level uses polynomial fitting to generate a rough trajectory, from which reference states and control inputs are derived for the second level. 
The second level formulates a model predictive control optimization problem in the Frenet frame, generating a trajectory that satisfies both kinematic feasibility and safety.
Experimental results on the F1TENTH platform show that our method outperforms the State-of-the-Art, achieving an 8.93\% higher overtaking success rate, allowing the maximum opponent speed, ensuring a smoother ego trajectory, and reducing 74.04\% computational time compared to the \emph{Predictive Spliner} method.
The code is available at: \url{https://github.com/ZJU-DDRX/FSDP}.
\end{abstract}


\section{Introduction} \label{sec:intro}
Autonomous racing serves as a high-intensity testbed for advancing general autonomous by pushing autonomy algorithms to their limits under extreme conditions \cite{betz_weneed_ar, ar_survey}. In recent years, competitions such as \gls{fsd}, the \gls{iac}, and the scaled-down F1TENTH races have gained popularity~\cite{f110, forzaeth}. While full-scale events like \gls{fsd} and \gls{iac} impose constraints on overtaking due to cost considerations, scaled autonomous racing, such as F1TENTH, enables unrestricted overtaking. Success in scaled autonomous racing depends on effective overtaking strategies, requiring autonomous vehicles to drive at peak speed and navigate around competitors. Unlike highway overtaking, where vehicle dynamics are stable and predictable, autonomous racing involves vehicles operating at the limits of grip, making overtaking a unique and complex challenge~\cite{zhang2024buzzracer}.

The methods for planning overtaking trajectories can be primarily divided into two approaches: end-to-end direct trajectory generation and traditional hierarchical planning\cite{ar_survey}. Although some end-to-end overtaking methods based on reinforcement learning have been validated in commercial vehicles~\cite{shi2024task}, simulations~\cite{trumpp2024racemop}, and scaled-down racing cars~\cite{song2021autonomous,ghignone2024tc}, in real-world racing environments, such as the \gls{A2RL}~\cite{inproceedings}, the high-speed collisions and costly repair expenses pose significant risks. As a result, current approaches in real-world applications still rely on more traditional methods involving perception, planning, and control. From a hierarchical perspective, the main challenge lies in planing a collision-free overtaking trajectory while considering predicting the opponent's trajectory at the friction limit.

Planning such a trajectory at high speeds involves three significant challenges: 
1) \textbf{Opponent Prediction:} The trajectory must account for the future behavior of the overtaken vehicle, as sudden lateral movements for opponent avoidance could result in loss of control and potential sliding;
2) \textbf{Overtaking Feasibility:} The planned trajectory must adhere to fundamental vehicle dynamics without collision, ensuring that the low-level controller can track it accurately;
3) \textbf{Real Time Requirements:} Overtaking opportunities are brief at high speeds, making the efficiency of the planning solution critical for success.

To address these challenges, this paper proposes the \gls{FSDP} for overtaking, which uses the \gls{sgp} to fit the opponent's behavior online. 
By combining Gaussian predictions with the vehicle's dynamics, \gls{FSDP} generates a collision-free and feasible overtaking trajectory with a bi-level \gls{qp} planner.
The main contributions of this paper are summarized as follows:
\begin{enumerate}[I]
    \item 
    \textbf{Sparse \gls{gp} based Prediction:} \gls{sgp} is utilized for real-time fitting of the opponent's trajectory and velocity, while its integration with K-means clustering enables efficient and accurate dataset updates. This approach reduces prediction time by 60.53\% and enhances prediction accuracy by 33.87\% compared to previous methods~\cite{baumann2024predictive}.
    \item 
    \textbf{Fast and Safe Overtaking Planning:} A bi-level \gls{qp} framework is used to efficiently generate a smooth and collision-free overtaking trajectory. 
    Both \gls{qp} optimizations incorporate opponent prediction information from \gls{sgp}.
    The first \gls{qp} provides a smooth initial trajectory via polynomial fitting. 
    Based on \gls{mpc}, the second \gls{qp} ensures kinematic feasibility and enforces safety using collision avoidance constraints in the Frenet frame. This enables \gls{FSDP} to achieve an 8.93\% higher overtaking success rate over~\cite{baumann2024predictive} with the maximum opponent speed.
    \item 
    \textbf{Open Source:} 
    The proposed \gls{FSDP} was tested and compared with \gls{sota} methods in simulations and on the F1TENTH car platform. 
    The complete code has been made open-source to enhance reproducibility~\cite{forzaeth}.
\end{enumerate}

\section{Related Work} \label{sec:Rela_work}
The existing overtaking planning methods in the context of autonomous racing can be mainly divided into two categories: sampling-based overtaking and optimization-based overtaking.

\subsection{Sampling Based Path-Planning}
Sampling-based planning methods primarily generate numerous candidate trajectories, followed by the design of a cost function to select the overtaking trajectory that meets the desired requirements. The candidate trajectory generation methods can be both real-time and offline. The \emph{Frenet Planner} \cite{frenetplanner} generates multiple trajectories online in the Frenet coordinate system. Then the cost function is designed to penalize deviations from the racing line and to avoid current opponents. A classic offline candidate trajectory generation method used in autonomous racing is called \gls{gbo} \cite{gbo}. This method generates candidate trajectories offline by defining nodes in a known map using a lattice and connecting the nodes with splines. Although sampling-based methods can efficiently generate overtaking trajectories, the choice of cost function and the number of sampling points can significantly impact the quality of trajectory generation. Additionally, the generated trajectories do not account for the future movement of opponents, which may lead to frequent changes in overtaking behavior and result in substantial computational requirements. 

\gls{FSDP} addresses these challenges by predicting the movement of the opponent car using Gaussian processes and incorporating this information into the \gls{qp} optimization problem
of the \gls{mpc}, significantly enhancing safety.

\subsection{Optimization-Based Overtaking} 
Optimization-based overtaking methods involve solving a real-time optimization problem with collision avoidance constraints.

\gls{mpcc} is an overtaking method that integrates both planning and control. Unlike methods that rely on a reference racing line, \gls{mpcc} aims to maximize the vehicle’s progress along the centerline, incorporating the opponent car as an obstacle within the optimization problem~\cite{mpcc}. It focuses solely on minimizing lap time while ensuring collision-free operation within the map. However, the performance of this method is highly sensitive to the parameters, and the controller also experiences a significant computational burden.
 
Another approach is to use polynomials as model constraints while treating the opponent as an obstacle in the optimization. A prominent example in racing is \emph{Spliner}~\cite{forzaeth}. This method fits a simple first-order polynomial in the Frenet coordinate system. Its advantages include high real-time performance and a high success rate of solving. \emph{Predictive Spliner}~\cite{baumann2024predictive} utilizes standard Gaussian processes to predict the trajectory and velocity of the opponent car. In the \emph{Spliner} optimization problem, the predicted collision intervals are considered to avoid collision, thereby achieving a higher overtaking success rate. This method has been successfully validated in the F1TENTH competition and is currently one of the \gls{sota} approaches. However, as the Gaussian dataset grows, both training and prediction times increase accordingly. Using only the most recent data to construct the dataset results in losing historical information about the opponent car. Although the sensitivity of the polynomial-based \emph{Spliner} parameters is significantly lower compared to the dynamics-based \gls{mpcc}, the generated trajectories may not be suitable for tracking by the low-level controller~\cite{becker2023model}.

\gls{FSDP} addresses these issues by employing sparse Gaussian processes to fit the trajectory and velocity of the opponent car and designing a distance function based on induced points to filter and update the data. Additionally, the \gls{mpc} utilizes a linearized kinematic model through differential flatness, enhancing both the feasibility of the trajectory and the computational efficiency of the solution.  

\subsection{Summary of Existing Overtaking Algorithms} 
\cref{tab:overtaking_algorithms} summarizes the comparison between the method proposed in this paper and \gls{sota} methods.
\begin{table} 
    \centering 
    \resizebox{\columnwidth}{!}{%
    \begin{tabular}{l|c|c|c|c}
    \toprule
    \textbf{Algorithm} & \textbf{Fast Opp. Pre} & \textbf{Kine. Feasibility} & \textbf{Param. Sensitivity} & \textbf{Compute} \\
    \midrule
    Frenet \cite{frenetplanner} & No & No & \textbf{Low} & \textbf{Low} \\
    \gls{gbo} \cite{gbo} & No & No & High & \textbf{Low} \\
    Spliner \cite{forzaeth} & No & No & \textbf{Low} & \textbf{Low} \\
    Pred. Spliner \cite{baumann2024predictive} & No & No & \textbf{Low} & Medium \\
    \gls{FSDP} \textbf{(ours)} & \textbf{Yes} & \textbf{Yes} & \textbf{Low} & \textbf{Low}  \\
    \bottomrule
    \end{tabular}%
    }
    \caption{Summary of existing overtaking strategies, denoting key characteristics of each method; \textit{Fast Opponent Prediction} considerations (desired yes); \textit{Kinematic model feasibility} is satisfied by the overtaking trajectory (desired yes); Susceptibility to \textit{Parameter Sensitivity} (desired low); Nominal Computational load (desired low). }
    \label{tab:overtaking_algorithms}
    \vspace{-2.0em}
\end{table}
\section{Methodology} 
\label{sec:methodology}
In this section, we provide a detailed illustration of how \gls{FSDP} operates. 
First, we use \gls{sgp} to predict the behavior of the opponent. 
Next, we demonstrate how to obtain an initial rough trajectory through polynomial fitting and differential flatness techniques. 
Finally, \gls{mpc} is employed to ensure the vehicle’s kinematic feasibility and enforce strict safety constraints.
The structure of \gls{FSDP} can be seen in Fig.~\ref{fig:graphical_abstract}.

\subsection{Sparse GP-Based Opponent Prediction} 
\label{ssec:res_ctrl_arch}

\subsubsection{Opponent Prediction with Sparse GP}
In the context of autonomous racing competitions, when a vehicle is in a trailing state, the perception system collects data on the position and speed of the opponent vehicle~\cite{forzaeth}. \emph{ Predictive Spliner}~\cite{baumann2024predictive} uses a standard \gls{gp} for fitting and prediction. Assume the dataset size is $N$, with input and output dimensions being $n_z$ and $n_d$, respectively. The computational complexities for calculating the mean and variance are $\mathcal{O}(n_d n_z N)$ and $\mathcal{O}(n_d n_z N^2)$. The proposed method uses \gls{sgp} with $M$ inducing points ($M \ll N$) to approximate the low-dimensional kernel matrix, mapping $\mathbb{R}^{N \times N}$ to $\mathbb{R}^{M \times M}$. The computational complexities for mean and variance become $\mathcal{O}(n_d n_z M)$ and $\mathcal{O}(n_d n_z M^2)$ \cite{liu2020gaussian}. Therefore, \gls{sgp} achieves fast fitting and prediction.

The choice of the sparse Gaussian method is based on \gls{VFE}~\cite{liu2020gaussian}. 
This method involves jointly inferring the inducing points and the hyperparameters by maximizing a lower bound of the marginal likelihood. Specifically, the approximate posterior is assumed to be $q(\bm{f},\bm{f}_z|\mathbf{Y})=p(\bm{f}|\bm{f}_z)\phi(\bm{f}_z)$, where $\bm{f}$ represents the latent function, $\bm{f}_z$ denotes the latent function at the inducing points $\mathbf{Z}_m$, $\mathbf{Z}$ is the input of the training data, and $\mathbf{Y}$ represents the noisy outputs. 
$\phi (\bm{f}_z)$ is a variational distribution over $\bm{f}_{z}$. This assumption enables a critical cancellation that results in a computationally tractable lower bound $\mathcal{F}_V$. After obtaining the optimal lower bound, the mean and covariance function of the approximate posterior at test point $z^{*}$ are obtained as
follows:
\begin{align}
    \mu^v(z^*) &= (\Sigma^{\omega})^{-1}K_{z^{*}\mathbf{Z}_m}(\mathcal{W})K_{\mathbf{Z}_m \mathbf{Z}}\mathbf{Y} \\
    \Sigma^v(z^*) &= K_{z^*z^*}-Q_{z^*z^*}+K_{z^{*}\mathbf{Z}_m}(\mathcal{W})K_{\mathbf{Z}_mz^{*}}
\end{align}
where $\mathcal{W} =(K_{\mathbf{Z}_m\mathbf{Z}_m}+(\Sigma^{\omega})^{-1}K_{\mathbf{Z}_m\mathbf{Z}}K_{\mathbf{Z}\mathbf{Z}_m})^{-1}$.

The travel distance $s_\text{opp}$, lateral error $d_\text{opp}$, and velocity $v_\text{opp}$ of the opponent car on the reference trajectory in the Frenet coordinate system are obtained through the perception system~\cite{forzaeth}. Using $s$ as the input, two sparse Gaussian models, denoted as $\mathcal{SGP}_d(s)$ and $\mathcal{SGP}_v(s)$, are constructed with \( d \) and \( v \) as the outputs, respectively.

\subsubsection{Data Selection}
In \emph{Predictive Spliner}, the \gls{gp} always uses the latest dataset for fitting and prediction, ignoring the information from old data. In contrast, our data selection method strategically constrains the dataset size $N_{\text{target}}$ while integrating both new observations and old data to achieve robust fitting of the preceding vehicle's trajectory. 

We construct two \gls{sgp} models with the input being the travel distance \( s \), and the outputs corresponding to the lateral error \( d \) and the speed \( v \), respectively. To facilitate the description of the data selection method, we will use the following notation: Given the training set \( \mathcal{D}_{\text{train}} = \{(\bm{x}_i, \bm{y}_i)\}_{i=1}^{N_\text{train}} \) and the incoming data \( \mathcal{D}_{\text{incoming}} = \{(\bm{x}_j, \bm{y}_j)\}_{j=1}^{N_\text{incoming}} \), where \( \bm{x} \) is the input feature vector and \( \bm{y} \) is the corresponding target value vector. Then, three methods are employed to conduct preliminary filtering of the data.

First, the \emph{Time-based Spatial Filter} calculates the spatial bins independently for each lap and selects the maximum timestamp within each spatial bin:
    \setlength\abovedisplayskip{2pt}
    \setlength\belowdisplayskip{2pt}
    \begin{equation}
        \mathcal{D}_{\text{filtered}} = \bigcup_{k=1}^{\lceil L/\Delta s \rceil} \left\{ \arg\max_{(\bm{x},y,t) \in \mathcal{S}_k} t \right\}
        \label{eq:time_filter}
    \end{equation}
    where $L$ is total trajectory length, $\Delta s$ is spatial bin width proportional to waypoint density, $\mathcal{S}_k$ is $k$-th spatial bin, and $t$ is the timestamp of data point.

Second, the \emph{Range Filter} ensures that data points are within a predefined range of values $[y_{\text{min}}, y_{\text{max}}]$ by removing physically implausible or sensor-faulty outliers to ensure data quality. 

Third, the \emph{Confidence Interval Filter} applies when $N_\text{train} > \frac{2}{3} N_{\text{target}}$, checking if the data points fall within the 95\% confidence interval of the original \gls{sgp}. This preserves data consistent with the current model's predictions when the dataset is sufficiently large, preventing noise corruption and maintaining model stability.

When using Leave-One-Out Variances with \gls{gp} or \gls{sgp} to describe the information richness of the data, it is necessary to solve the inverse matrix of the new kernel at each step. The proposed method, which utilizes inducing points $\mathbf{Z}_m$ to describe the predictive distance of different points, significantly improves the real-time performance.
A smaller distance indicates that the information contained in the two points is more similar, as shown in the following equation:
\begin{equation}
    \sigma_s^2(\bm{x}_j) = k(\bm{x}_j, \bm{x}_j) - \bm{k}_{\bm{x}_j\mathbf{Z}_m} \mathbf{K}_{\mathbf{Z}_m\mathbf{Z}_m}^{-1} \bm{k}_{\mathbf{Z}_m\bm{x}_j} + \Sigma^{\omega}
    \label{eq:sparse_var}
\end{equation}
where $k(\cdot,\cdot)$ is kernel function and $\Sigma^{\omega}$ is observation noise variance.

After data filtering, new data can only be added to the dataset if its predictive distance exceeds the average distance of the training data.
This step aims to ensure that the dataset contains information-rich points, as described by the following equation:
\begin{equation}
    \setlength\abovedisplayskip{2pt}
    \setlength\belowdisplayskip{2pt}
    \sigma^2_s(\bm{x}_j) > \frac{1}{N_\text{train}}\sum_{i=1}^{N_\text{train}} \sigma_s^2(\bm{x}_i)
\end{equation}
where $\sigma_s^2(\bm{x}_j)$ is predictive variance at $\bm{x}_j$ from Eq.~\eqref{eq:sparse_var}.

The pruning process is activated only if the merged dataset $D_{\text {merged }}=D_{\text {train }} \cup D_{\text {filtered }}$ exceeds the preset size $N_{\text{target}}$. First, K-means clustering is performed using the \gls{sgp}'s inducing points $\mathbf{Z}_m$ as initial centroids, partitioning $D_{\text {merged }}$ into $K$ clusters (where $K$ equals the number of inducing points $M$). The clustering objective is:
\begin{equation}
    \setlength\abovedisplayskip{2pt}
    \setlength\belowdisplayskip{2pt}
\min _{\left\{c_k\right\}} \sum_{k=1}^K \sum_{x \in c_k}\left\|x-z_{m,k}\right\|^2
\end{equation}
where $z_{m,k}$ represents the k-th inducing point.

Next, the mean predictive distance $\bar{\sigma}^2_{s,k}$ within each cluster $C_k$ is calculated. Data points with variances below $\bar{\sigma}^2_{s,k}$ are removed. If the global dataset size remains larger than $N_{\text {target }}$ after this step, additional points are pruned proportionally from each cluster until the target size is reached.

\begin{algorithm}

\caption{Inducing-Guided Data Selection}
\label{algo:data_sel}
\begin{algorithmic}  [1]
\State \textbf{Input:} $\mathcal{D}_{\text{train}}$, $\mathcal{D}_{\text{incoming}}$\quad\quad\textbf{Output:} $\mathcal{D}_{\text{pruned}}$  
\State \textbf{Filtering:} 
\For{each lap}
    \State Sort $\mathcal{D}$ by $\Delta s$, keep latest point in each bin 
\EndFor
\State $\mathcal{D}_{\text{filtered}} = \{\bm{x}_j, y_j \in \mathcal{D}_{\text{incoming}} \mid y_j \in [y_\text{min}, y_\text{max}]\}$
\If{$|\mathcal{D}_{\text{train}}| > \frac{2}{3}N_\text{target}$}
   \State Retain $y \in [\mu(x) \pm 1.96\sigma^2(x)]$ for $(x, y) \in \mathcal{D}_{\text{filtered}}$
\EndIf
\State \textbf{Adding New Data:} 
\State Keep $\sigma_s^2(x) > \frac{1}{|\mathcal{D}_{\text{train}}|}\sum \sigma_s^2(\mathcal{D}_{\text{train}})$ for $x \in \mathcal{D}_{\text{filtered}}$ 
\State $\mathcal{D}_{\text{merged}} \gets \mathcal{D}_{\text{train}} \cup \mathcal{D}_{\text{filtered}}$
\State \textbf{Pruning:} 
\If{$|\mathcal{D}_{\text{merged}}| \geq N_\text{target}$}  
    \State Cluster $\mathcal{D}_{\text{merged}}$ via K-means($\mathbf{Z}_m$) 
    \For{each cluster $C_k$}
        \State Prune $\bm{x}$ where $\sigma_s^2(\bm{x}) < \frac{1}{|C_k|}\sum \sigma_s^2(C_k)$
        \State Keep top $\lfloor N_\text{target} \cdot \frac{|C_k|}{|\mathcal{D}_{\text{merged}}|} \rfloor$ points
    \EndFor
\EndIf
\end{algorithmic}
\end{algorithm}

\subsubsection{Collision Prediction} 
Let \( s_\text{ego}(k) \) and \( s_\text{opp}(k) \) represent the positions of the ego and opponent vehicles at time \( k \). The time interval and prediction horizon are defined as $\Delta t$ and $\mathcal{T}_N$. The forward simulation of the vehicles is shown below:
\begin{equation}
\begin{aligned}
 s_\text{ego}(k+1) &= s_\text{ego}(k) + v_\text{ego}(k) \Delta t + \frac{1}{2} a_\text{ego}(k) \Delta t^2 \\
 v_\text{ego}(k+1) &= v_\text{ego}(k) + a_\text{ego}(k) \Delta t \\
 s_\text{opp}(k+1) &= s_\text{opp}(k) + v_\text{opp}(k) \Delta t \\
 v_\text{opp}(k+1) &= \mathcal{SGP}_v(s_\text{opp}(k+1))
\end{aligned}
\end{equation}

Within the prediction horizon $T_N$, if the difference in travel distance between the two vehicles is less than the collision threshold \( s_c \), the position of the ego vehicle at the corresponding point \( s_\text{ego}(k) \) is identified as the start of the collision interval \(c_\text{start}\). Subsequently, if the difference in travel distance exceeds the collision threshold \( s_c \), the point \(c_\text{end}\)  is marked as the end of the collision interval. After obtaining the collision interval, the lateral position of the opponent car can be represented using the \gls{sgp} model based on the vehicle's position. 
\begin{align}
d_\text{opp} = \mathcal{SGP}_d(s), \quad s \in [c_\text{start}, c_\text{end}]
\end{align}
The predicted collision interval provides a safe constraint for generating the overtaking trajectory in the subsequent steps.

\subsection{Efficient QP-Based Initial Trajectory Generation}
\begin{figure}
    \centering
    \includegraphics[width=0.9\columnwidth]{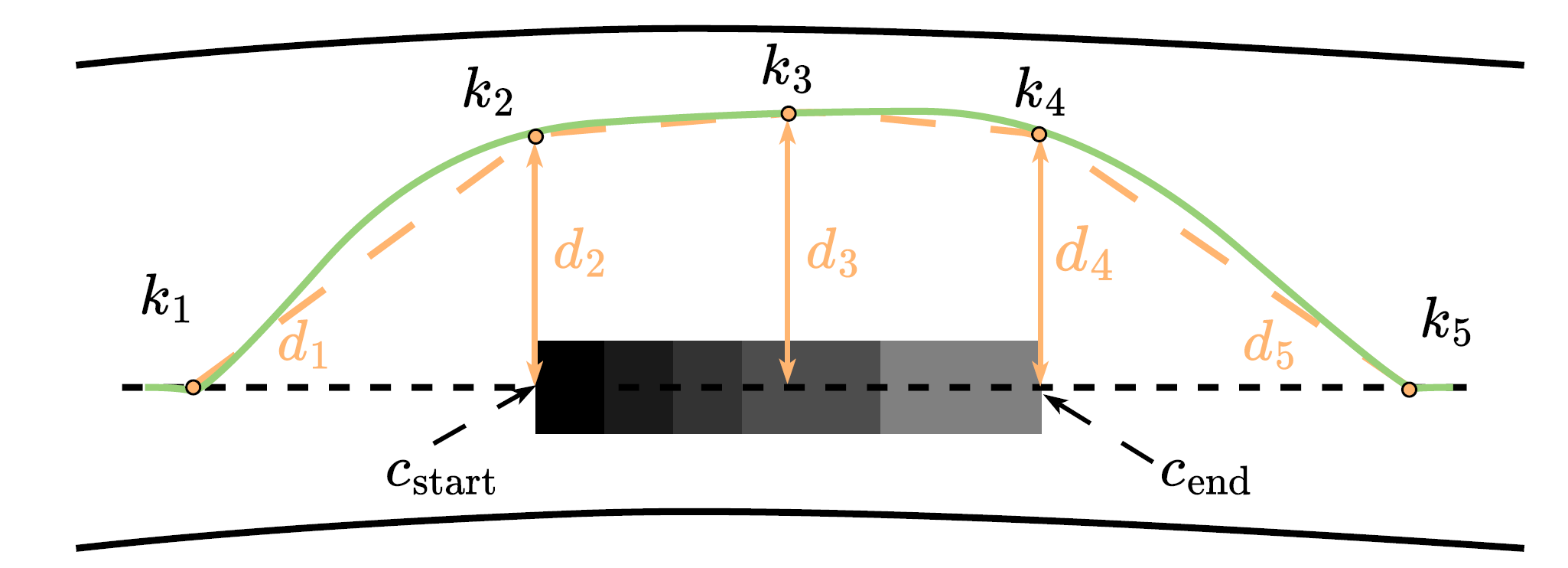}
    \caption{Schematic diagram of the initial collision-free trajectory. The black region represents the predicted positions of the opponent using \gls{sgp}. The black dashed line denotes the racing line, while the yellow dashed line illustrates the rough evasion trajectory obtained through key point interpolation. The green trajectory is the smoothed avoidance trajectory obtained from \gls{qp}.}
    \label{fig:sample_fit}
    \vspace{-2.0em}
\end{figure}

In this section, a collision-free initial trajectory is rapidly generated by constructing a \gls{qp} fitting problem. Based on the predicted positions of opponents from \gls{sgp}, several key points are identified in the Frenet frame, such as $k_1$, $k_2$, $k_3$, $k_4$, and $k_5$, as shown in Fig.~\ref{fig:sample_fit}. $k_1$ and $k_5$ represent the start and end points of the planned trajectory, with the corresponding $d=0$. $k_2$, $k_3$, and $k_4$ represent the start, middle, and end points of the predicted opponent trajectroy. The $d$ values for $k_2$, $k_3$, and $k_4$ are then determined, ensuring that the points maintain a distance greater than the safety threshold from the corresponding predicted opponent and do not exceed the track boundaries.

After determining the $d$ coordinates of these key points, linear interpolation is performed to estimate the intermediate $d$ coordinates. These discrete points are used to generate an initial, collision-free trajectory $P^\star(t)$. A \gls{qp} problem is then formulated to refine this trajectory by fitting it with a polynomial, ensuring a smoother trajectory.

The vehicle's position is described by the polynomial $P(t) = \bm{k}^\top\bm{\beta(t)}$, where $t \in [0, T]$, $\bm{k}\in\mathbb{R}^{2n}$ is the coefficient vector, and $\bm{\beta(t)} = \begin{bmatrix} 1 & t & t^2 & \cdots & t^{2n-1} \end{bmatrix}^\top$ is the natural basis.
The trajectory is a fifth-degree polynomial since $n=3$. The trajectory fitting problem can be formulated as a \gls{qp} problem as follows:
\begin{equation}
  \begin{split}
    &\min \,\, {\int_{0}^{T}\left\|P(t)-P^\star(t)\right\|^2dt}\\
    &s.t.\quad  \begin{array}{lc}
      P^{(j)}(T)=P^{\star(j)}(T) \\
      P^{(j)}(0)=P^{\star(j)}(0),j\in\{0,1\}
    \end{array}
  \end{split}\label{candidate_opti}
\end{equation}
where the constraints include position and velocity constraints at the start and end points. The problem (\ref{candidate_opti}) is solved using OSQP \cite{stellato2020osqp}.

\subsection{Linearized Reference States Extraction Based on Differential Flatness}
After the trajectory represented by the fifth-degree polynomial is obtained, the differential flatness property is used to find the linearized reference states, facilitating the lower-layer \gls{mpc} optimization. For the kinematic bicycle model in Cartesian coordinates frame, the state vector $\mathbf{x}_c = [x, y, v, \theta]^\top$, the input vector $\mathbf{u}_c = [a_t, \delta]^\top$ are defined. Where $[x, y]^\top$ is the position at the center of the rear wheel, $v$ is the longitudinal velocity, $a_t$ is the longitude acceleration, and $\delta$ is the steering angle. Based on the characteristics of differential flatness, all information is expressed through $x$ and $y$ as follows:
\begin{subnumcases} {\label{df_eq}}
  v = \sqrt{\dot{x}^2+\dot{y}^2}\label{df_v}\\
  \theta = \tan^{-1}(\dot{y}/\dot{x})\label{df_theta}\\
  a_t = (\dot{x}\ddot{x}+\dot{y}\ddot{y})/\sqrt{\dot{x}^2+\dot{y}^2}\label{df_a}\\
  \delta = \tan^{-1}((\dot{x}\ddot{y}-\dot{y}\ddot{x})L/(\dot{x}^2+\dot{y}^2)^{\frac{3}{2}})\label{df_delta}
\end{subnumcases}
where $L$ is the length of the wheelbase. Thus, the flat outputs $[x,y]^\top$ and their finite derivatives describe the arbitrary state and input information of the vehicle \cite{han2023efficient}. After selecting the reference points along the trajectory and coordinate transformation, the corresponding states $\mathbf{X}_\text{ref}$ and control inputs $\mathbf{U}_\text{ref}$ in the Frenet frame can be obtained using differential flatness. These reference points can then serve as the linearized reference points for \gls{mpc} in the lower-layer planner.

\subsection{MPC-Based Optimization for Kinematic Feasibility and Safety Guarantee}
This section demonstrates the optimization of initial reference trajectory. 
To satisfy the vehicle's kinematic dynamics and ensure safety, \gls{mpc} is employed for optimization, and the entire optimization problem is formulated in the Frenet frame to ensure real-time performance.
The vehicle's kinematic dynamics are expressed as follows:
\begin{equation}
    \begin{split}
    \dot{s} &= \frac{v\cos{\theta}}{1 - \kappa_r(s) n}, \; \dot{n} = v\sin{\theta}, \\
    \dot{\theta} &= \frac{v\tan{\delta}}{L} - \kappa_r(s) \frac{v\cos{\theta}}{1 - \kappa_r(s) n},
    \end{split}
\end{equation}
where $s$ denotes the distance traveled along the racing line, \( \theta \) represents the heading angle error, \( \kappa_{r} \) represents the reference curvature, and \( n \) is the lateral deviation. 
Suppose $\bm{x} = [s, n, \theta]^\top$ and $\bm{u} = [v, \delta]^\top$, where \( n_x = 3 \) and \( n_u = 2 \) are defined.

Assume the state sequence over the \gls{mpc} prediction horizon $N$ is denoted by $\mathbf{X} = [\bm{x}_1^\top, \dots, \bm{x}_N^\top]^\top \in \mathbb{R}^{N \cdot n_x}$, and $\mathbf{U} = [\bm{u}_0^\top, \dots, \bm{u}_{N-1}^\top]^\top\in \mathbb{R}^{N \cdot n_u}$ denotes the control inputs sequence.
In addition, we define an integrated state $\mathbf{X}^* = [\mathbf{X}^\top - \mathbf{X}_\text{ref}^\top, \mathbf{X}^\top, \mathbf{X}^\top - \mathbf{X}_\text{pre}^\top]^\top$, where $\mathbf{X}_\text{ref}$ denotes the reference states sequence derived from differential flatness, and $\mathbf{X}_\text{pre} = [\bm{x}_0^\top, \dots, \bm{x}_{N-1}^\top]^\top$. The optimization problem is formulated as:
\begin{subequations}
\begin{align}
    \min_{\mathbf{X}, \, \mathbf{U}} \, & \| \mathbf{X}^{*} \|_\mathbf{Q}^2 + \| \mathbf{U} - \mathbf{U}_\text{ref}\|_\mathbf{R}^2 \label{eq:obj} \\
    \text{s.t.} ~& \bm{x}_k = \mathbf{A}_k \bm{x}_{k - 1} + \mathbf{B}_k \bm{u}_{k - 1}, k = 1, 2, \dots, N, \label{eq:dynamics} \\
    & l_b(s_k^\text{ref}) \leq \bm{x}_k(1) \leq u_b(s_k^\text{ref}), k = 1, 2, \dots, N, \label{eq:ca_cons} \\
    & \bm{u}_\text{min} \leq \bm{u}_k \leq \bm{u}_\text{max}, \Delta \bm{u}_\text{min} \leq \Delta \bm{u}_k \leq \Delta \bm{u}_\text{max}, \label{eq:physical_limits}
\end{align}
\label{eq:mpc_problem}
\end{subequations}

\vspace{-1em}
\noindent where $\Delta \bm{u}_k = \bm{u}_k - \bm{u}_{k - 1}$, and $\mathbf{A}_k$ and $\mathbf{B}_k$ are the linearized system matrices evaluated at the reference points $\bm{x}_{k, \text{ref}}$ and $\bm{u}_{k - 1, \text{ref}}$. $\mathbf{U}_\text{ref}$ represents the reference control sequence from differential flatness. $\mathbf{Q} = \operatorname{blkdiag}(\mathbf{Q}_1,\mathbf{Q}_2,\mathbf{Q}_3)$ and $\mathbf{R}$ are weight matrices. The term associated with $\mathbf{Q}_1$ aims to follow the reference trajectory closely. 
Among the diagonal elements of $\mathbf{Q}_1,\mathbf{Q}_2$, only those associated with the lateral error variable are positive to ensure the trajectory closely follows the racing line while maintaining smoothness.

Constraint~\eqref{eq:ca_cons} imposes safety on the states by confining the lateral deviation $n$ in a reasonable scope to avoid collisions. $l_b$ and $u_b$ represent the boundary of the lateral deviation $n_k$ at the $s_k^\text{ref}$. 
Since the sparse Gaussian model predicts the collision interval, the left and right boundaries of the lateral error take into account the future motion of the opponent car. 
For example, considering the left boundary $l_b$:
\begin{equation}
    l_b = \begin{cases}
         d_\text{opp}, & \text{if}\, c_\text{start} < s_k^{\text{ref}} < c_\text{end} \\
        d_{l}^{\text{ref}}, & \text{otherwise}
    \end{cases}
\end{equation}
where \( d_\text{opp} \) represents the predicted position of the opponent car and \( d_{l}^{\text{ref}} \) is the left boundary of the track.
The future motion of the opponent vehicle is incorporated into the \gls{mpc}-based \gls{qp} problem~\eqref{eq:mpc_problem}, thereby enhancing both safety.

\section{Experimental Results} 
\label{sec:exp_results}
This section evaluates the proposed \gls{FSDP} method in comparison with the \gls{sota} overtaking methods listed in \cref{tab:overtaking_algorithms}, focusing on overtaking success rate and trajectory feasibility. 

\subsection{Experimental Setup}
The evaluation will be divided into two parts: simulation and physical verification.
The simulation scene uses the default F1TENTH simulator map, and the official F1TENTH simulator~\cite{f110} is used to model the behavior of a 1:10 scale racing car in real-world scenarios. The CPU used for the simulations  is an AMD Ryzen 7 5800H running at 3.2 GHz.
Additionally, the physical experiment was conducted on an F1TENTH racecar based on~\cite{forzaeth}, using an Intel NUC as the onboard computer. The sensor suite includes a Hokuyo UST-10LX LiDAR with a maximum scan frequency of 40 Hz. Odometry is provided by the \gls{VESC}. Sensor fusion combines data from the LiDAR, the IMU embedded in the \gls{VESC}, and odometry, utilizing the Cartographer to localize the vehicle and obtain its state.

During the experiment, both the ego car and the opponent car will follow the global minimum curvature racing line~\cite{gb_optimizer}.
The trajectory tracking controller uses a tire-dynamics-based PP controller, with tire parameters identified through an on-track system identification method~\cite{dikici2025learning}. To better evaluate the overtaking ability of different algorithms, we define the speed-scaler \bm{$\mathcal{S}_\text{max}$} $= \frac{v_\text{opp}}{v_\text{ego}} \in [0, 1]$. The ego vehicle runs at its maximum trackable reference speed, while the opponent moves at the highest speed that still allows overtaking. The speed-scaler reflects their relative speed ratio and serves as an indicator of overtaking difficulty.
In the experiments, we evaluate different algorithms by progressively reducing the opponent's speed until the ego vehicle is able to successfully complete the overtaking maneuver.
The metrics used to evaluate the algorithm's performance are as follows:
\begin{itemize}
  \item $\bm{\mathcal{R}_\text{ot/c}}$: The number of successful overtakes $\mathcal{N}_\text{ot}$ and crashes $\mathcal{N}_\text{c}$ were logged to compute the overtaking success rate $\mathcal{R}_\text{ot/c} = \frac{\mathcal{N}_\text{ot}}{\mathcal{N}_\text{ot} + \mathcal{N}_\text{c}}$.
  \item $\bm{l}$\label{len}: The length of overtaking trajectory.
  \item $\bm{T}$\label{T}: The time taken to complete the overtaking maneuver and return to the race line.
  \item $\bm{\bar{\dot{a}}}$\label{bar_a}: The average change rate in acceleration during overtaking.
  \item $\bm{\bar{{\dot{\delta}}} }$\label{dot_delta}: The average rate of change of the steering angle during overtaking.
\end{itemize}
$\mathcal{R}_\text{ot/c}$ is used to indicate the success rate of the overtaking algorithms. $l$ and $T$ represent the execution efficiency of the overtaking trajectory, while ${\bar{\dot{a}}}$ and $\bar{{\dot{\delta}}}$ describe the stability of the vehicle when tracking the overtaking trajectory. The planner is only activated when an obstacle is detected; otherwise, it remains in racing mode~\cite{forzaeth}.

\subsection{Simulation Results}
\subsubsection{Prediction}
In the simulation, a virtual opponent was generated with Gaussian noise to simulate opponent vehicles. After three laps, the proposed data selection method was compared with the \emph{Predictive Spliner} method. The latter utilizes only the latest data for path prediction, while the proposed method integrates both new and old data, thereby better handling noise and changes in the opponent's behavior.

\begin{figure}
    \centering
    \includegraphics[width=0.45\textwidth]{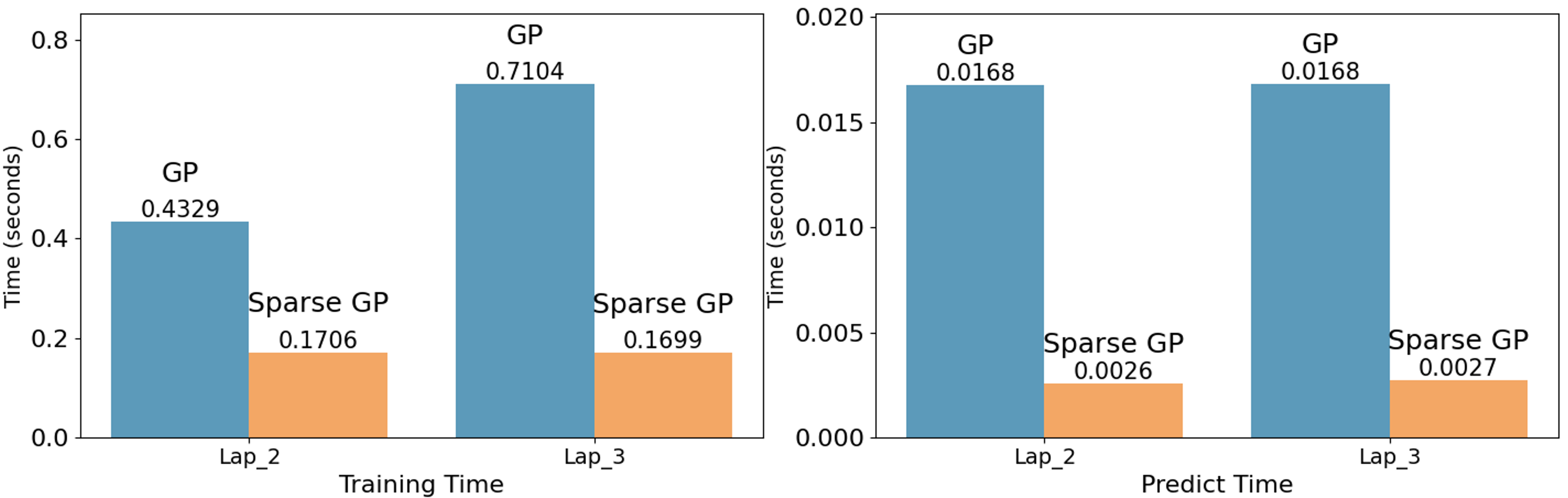}
    \caption{Training and prediction time comparison of the original \gls{gp} and the proposed \gls{sgp}, evaluated on the simulation environment.}
    \label{fig:sgp_time}
    \vspace{-1.0em}
\end{figure}

The dataset is limited to 400 samples to prevent it from growing indefinitely. As illustrated in \cref{fig:graphical_abstract}, the position and velocity of the opponent vehicle are predicted using \gls{sgp}. Taking the position prediction as an example, the training and prediction times of \gls{sgp} are reduced by 60.60\% and 83.93\%, respectively, compared to \gls{gp}, highlighting the superior real-time performance of \gls{sgp}, as shown in \cref{fig:sgp_time}.

\begin{figure}
    \centering
    \includegraphics[width=0.45\textwidth]{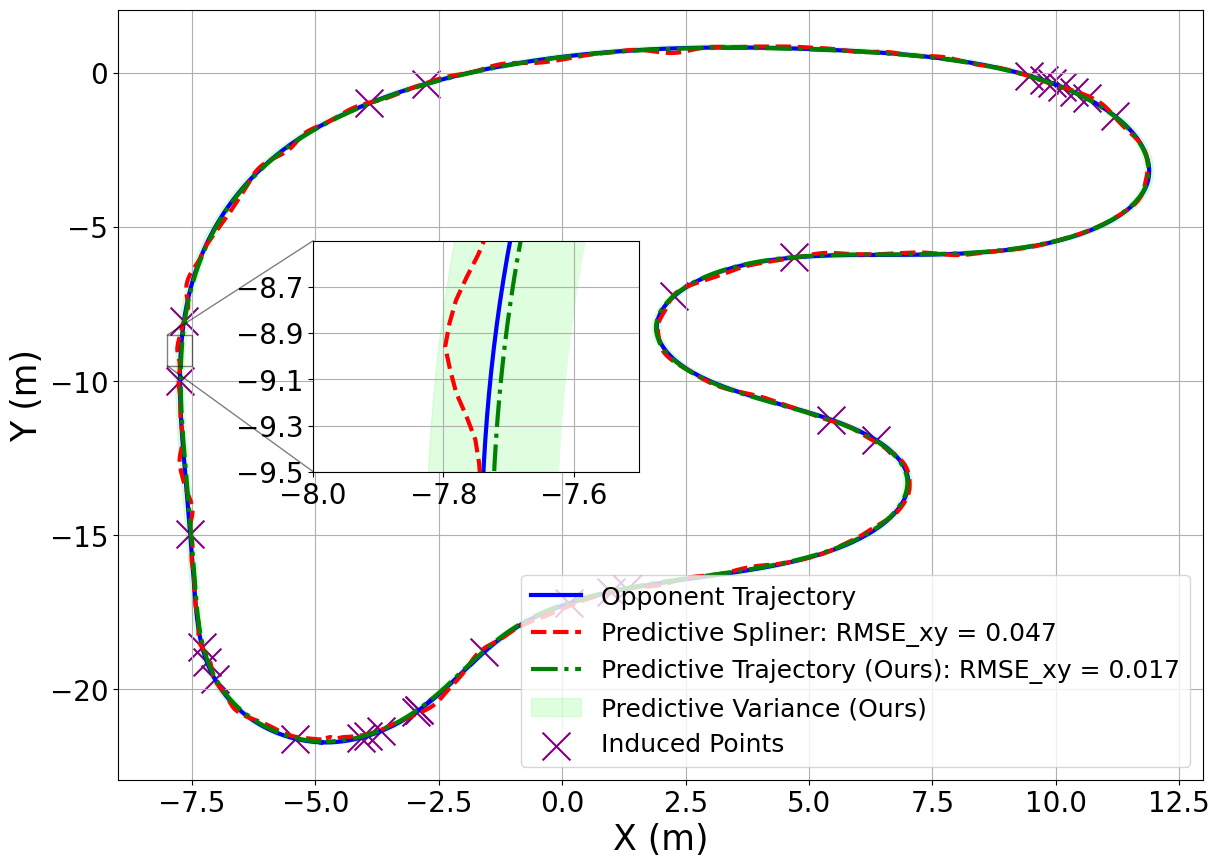}
    \caption{Predictive trajectory comparison of the baseline \emph{Predictive Spliner} \cite{baumann2024predictive} compared to the proposed \gls{FSDP} prediction. The opponent's actual trajectory is depicted in blue; the baseline \emph{Predictive Spliner} \cite{baumann2024predictive} in red; the proposed \gls{FSDP} solution in green with the green shaded variance, as well as the induced points of the \gls{sgp} in violet.}
    \label{fig:sgp_pre_traj}
    \vspace{-2.0em}
\end{figure}

In terms of updates to the \gls{gp} dataset, as shown in \cref{fig:sgp_pre_traj}, the proposed selection method reduces the RMSE of the predicted trajectory by 63.83\% compared to the \emph{Predict Spliner} method, making it more closely aligned with the actual movement of the virtual opponent. The \emph{Predict Spliner} method, relying solely on the latest data, is more vulnerable to noise interference and data distortion.

\begin{table}
    \centering
    \resizebox{0.45\textwidth}{!}
    {%
    \begin{tabular}{l|ccccc}
    \toprule
    \textbf{Overtaking Planner}
    &\bm{$\mathcal{R}_{ot/c}[\%]$} & \bm{$l[m]$}
    &\bm{$T[s]$} & \bm{${\bar{\dot{a}}}[m/s^3]$}
    &\bm{$\bar{{\dot{\delta}}}[rad/s]$}\\
    \midrule
    Frenet    & 57.14 & \textbf{13.53} & \textbf{3.86} & 237.18 & 0.71 \\
    GBO  & 57.14 & 15.96   & 5.04 & 443.12   & 0.62\\
    Spliner   & 72.72 & 22.82   & 5.35 & 70.13   & 0.55 \\
    Pred. Spliner  & 81.81 & 23.15 & 5.23 & 93.64 & 0.50 \\
    FSDP \textbf{(ours)}  & \textbf{90.90} & 22.67   & 4.85 & \textbf{42.72} & \textbf{0.40}\\
    \bottomrule
    \end{tabular}%
    }
    \caption{Overtaking performance of each evaluated planner in the simulation.}
    \label{tab:sim_res}
\end{table}
\subsubsection{Overtake}
\begin{figure*}
    \includegraphics[width=\textwidth]{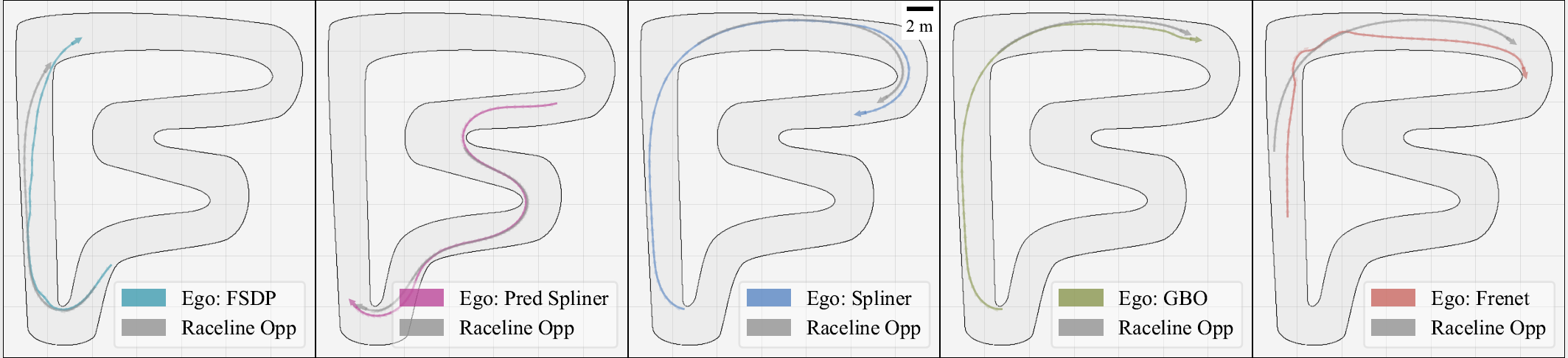}
    \caption{Comparison of our approach with \gls{sota} in the simulation of an overtaking maneuver. Each subplot illustrates the trajectories of the ego vehicle (colored path) and the opponent (gray path) during the maneuver.
    The speed scalar \bm{$\mathcal{S}_\text{max}$} for each method is as: FSDP: 53.8\%, Pred. Spliner: 53.8\%, Spliner: 47.5\%, GBO: 30\%, Frenet: 30\%.}
    \label{fig:qualitative_ots_sim}
    \vspace{-1.0em}
\end{figure*}
As shown in \cref{tab:sim_res}, the overtaking success rate of \gls{FSDP} reached 90.90\%, surpassing the performance of the listed \gls{sota}. For the same \bm{$\mathcal{S}_\text{max}$}, the overtaking success rate of \gls{FSDP} is approximately 9.09\% higher than that of the \emph{Predictive Spliner}. The average speed during overtaking is $v_{o} = l/T$. For FSDP, its $v_o = 4.67 \, \si[per-mode=symbol]{\metre\per\second}$ the highest among SOTA methods, indicating that the trajectories it plans have higher execution efficiency.
 Due to the absence of dynamic opponent prediction, methods such as \emph{{Frenet}}, \emph{\gls{gbo}}, and \emph{Spliner} show lower overtaking success rates under lower values of $\mathcal{S}_\text{max}$. Regarding trajectory quality, \gls{FSDP} exhibits smaller values for ${\bar{\dot{a}}}$ and $\bar{{\dot{\delta}}}$ in simulation tests. Compared to the \emph{Predictive Spliner}, these two metrics have decreased by 54.37\% and 20.0\%, respectively. This improvement can be attributed to the fact that \gls{FSDP} simultaneously considers both the smoothness of the trajectory and the vehicle's kinematic model. These results indicate that \gls{FSDP} not only achieves high success rates but also ensures high performance in terms of trajectory quality. \cref{fig:qualitative_ots_sim} visualizes the overtaking trajectories of all methods in simulation. 

\subsection{Physical Results}
\subsubsection{Prediction}
\begin{figure}
    \centering
    \includegraphics[width=0.45\textwidth]{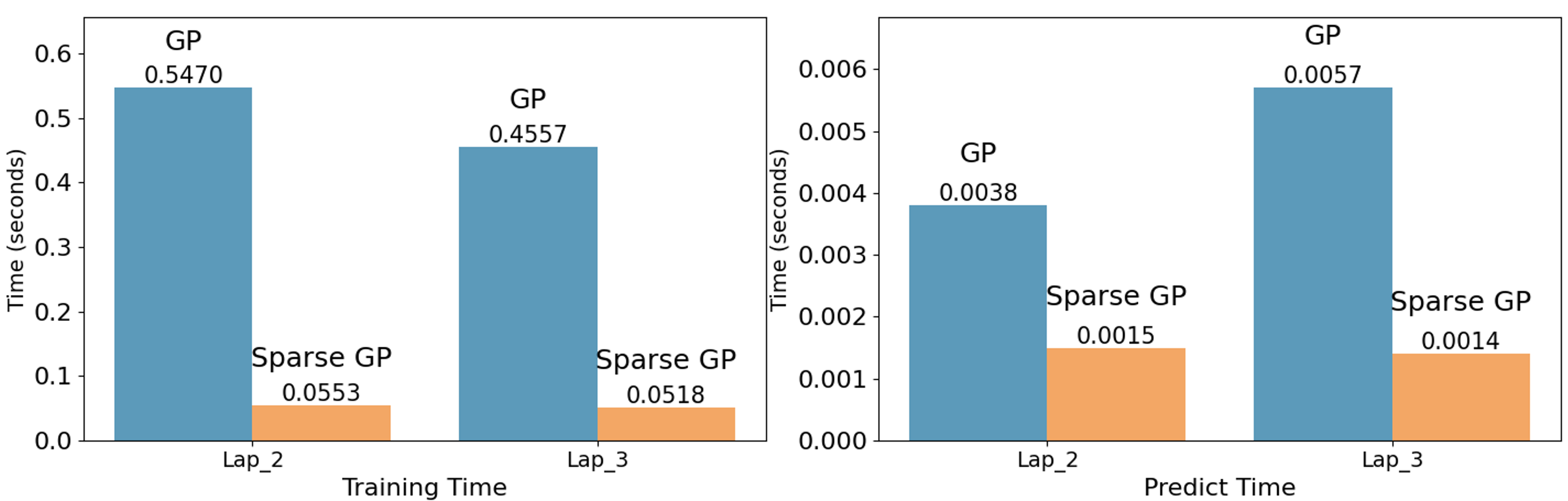}
    \label{Training And Predict Time Comparison in Real Car}
    \caption{Training and prediction time comparison of the original \gls{gp} and the proposed \gls{sgp}, evaluated on the on-board compute unit of the physical robot.}
    \label{fig:sgp_real_comp_time}
    \vspace{-1.0em}
\end{figure}

\begin{figure}
    \centering
    \includegraphics[width=0.45\textwidth]{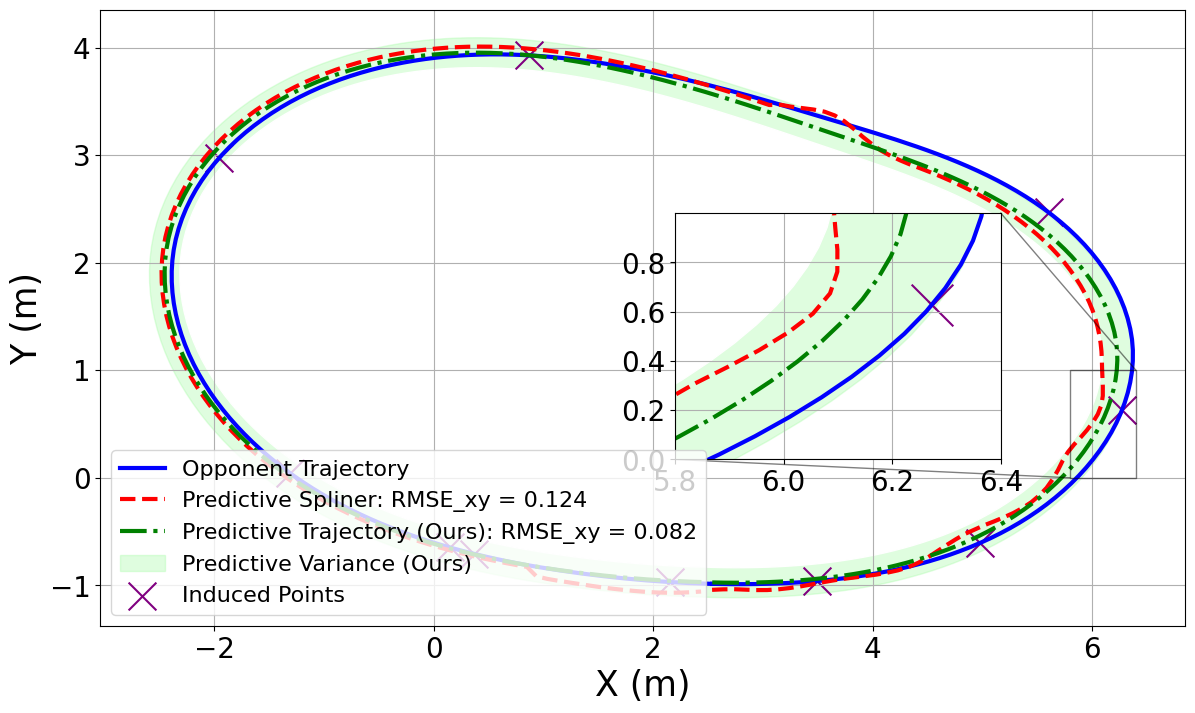}
    \label{Predictive Trajectory Comparison in Real Car}
    \caption{Predictive trajectory comparison of the baseline \emph{Predictive Spliner} \cite{baumann2024predictive} compared to the proposed \gls{FSDP} prediction in the real map.}
    \label{fig:sgp_real_pre_traj}
\end{figure}

In the real vehicle experiment, the same process as in the simulation was followed, where a real vehicle replaced the virtual opponent. After three laps, the data-filtering results of our method were compared with those of the \emph{Predictive Spliner} method. With \gls{sgp}, reductions in training time by 88.63\%, prediction time by 60.53\%, and a decrease in RMSE of the predicted trajectory by 33.87\% were achieved after the data selection. The specific results are presented in Fig.~\ref{fig:sgp_real_comp_time} and Fig.~\ref{fig:sgp_real_pre_traj}. Results demonstrate the real-time performance of \gls{sgp}, as well as the method's ability to more accurately predict the opponent's behavior by integrating both new and historical data.

\begin{table}
    \vspace{-2.0em}
    \centering
    \vspace{4.5mm}
    \resizebox{0.45\textwidth}{!}
    {%
    \begin{tabular}{l|ccccc}
    \toprule
    \textbf{Overtaking Planner}&\bm{$\mathcal{R}_{ot/c}[\%]$} & \bm{$l[m]$}
    &\bm{$T[s]$} & \bm{${\bar{\dot{a}}}[m/s^3]$}
    &\bm{$\bar{{\dot{\delta}}}[rad/s]$}\\
    \midrule
    Frenet    & 62.50 & 5.82    & 3.81 & 1082.26   & 0.96 \\
    GBO   & 55.00 & 8.89   & 5.02 & 916.15   & 1.08\\
    Spliner   & 60.00 & 7.18   & 3.02 & 773.54   & 0.81 \\
    Pred. Spliner & 62.50 & 5.56   & \textbf{2.00} & 824.27 & \textbf{0.68}\\
    FSDP \textbf{(ours)}  & \textbf{71.43} & \textbf{5.50}   & 2.02 & \textbf{218.15} & 0.70 \\
    \bottomrule
    \end{tabular}%
    }
    \caption{Overtaking performance of each evaluated planner on a 1:10 scale physical autonomous racing platform.}
    \label{tab:phys_res}
    \vspace{-2.0em}
\end{table}

\subsubsection{Overtake}
\begin{figure*}
    \includegraphics[width=\textwidth]{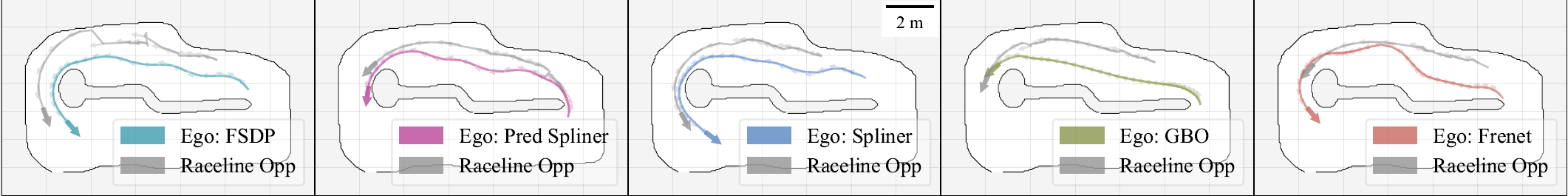}
    \caption{Comparison of our approach with \gls{sota} in the real experiment of an overtaking maneuver. Each subplot illustrates the trajectories of the ego vehicle (colored path) and the opponent (gray path) during the maneuver.
    The speed scalar \bm{$\mathcal{S}_\text{max}$} for each method is as: \gls{FSDP}: 60\%, Pred. Spliner: 55\%, Spliner: 45\%, GBO: 30\%, Frenet: 30\%.}
    \label{fig:qualitative_ots_phy}
    \vspace{-1.0em}
\end{figure*}
\begin{figure}
    \centering
    \includegraphics[width=0.38\textwidth]{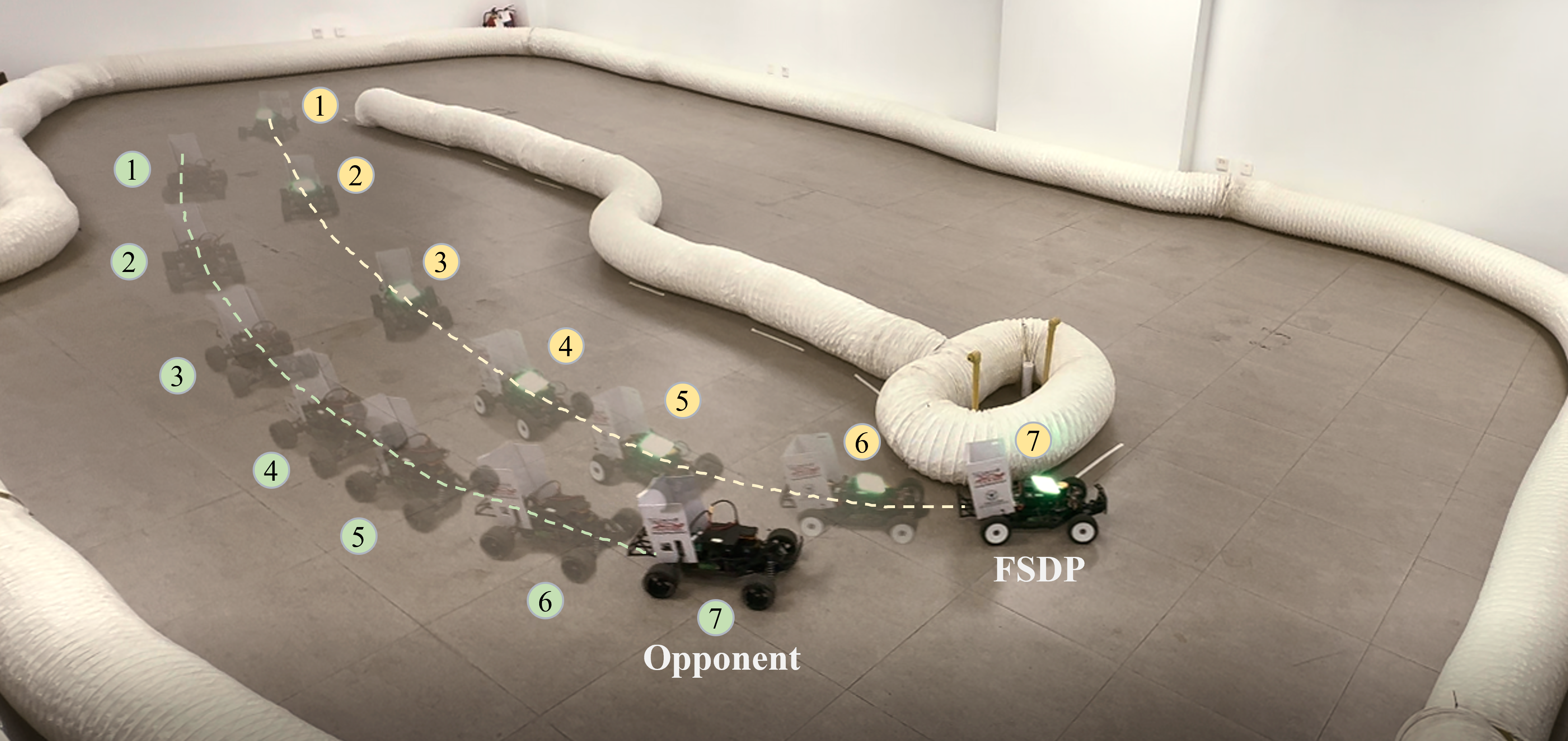}
    \caption{FSDP's Overtaking Screenshot.}
    \label{fig:FSDPovertake}
    \vspace{-2.0em}
\end{figure}

As shown in \cref{tab:phys_res}, compared to \gls{sota}, \gls{FSDP} still demonstrates advantages in trajectory execution efficiency and smoothness. In comparison to the \emph{Predictive Spliner}, the value of 
${\bar{\dot{a}}}$ in \gls{FSDP} decreased by 73.53\%. With the highest \bm{$\mathcal{S}_\text{max}$}, \gls{FSDP} achieved an overtaking success rate of 71.43\% on the real vehicle, representing the best performance among the \gls{sota} methods. Compared to the \emph{Predictive Spliner}, the success rate of \gls{FSDP} is 8.93\% higher, indicating improved safety. This result indicates that the trajectory planned by \gls{FSDP} offers higher safety and smoothness on real racetracks. 
\cref{fig:qualitative_ots_phy} visualizes the overtaking trajectories of all methods in the real vehicle experiments, and \cref{fig:FSDPovertake} shows \gls{FSDP}'s overtaking screenshot.

\subsection{Compute}
\begin{table}
    \centering 
    \resizebox{\columnwidth}{!}{%
    \begin{tabular}{l|cc|cc}
    \toprule
    \textbf{Overtaking Planner}& \multicolumn{2}{c|}{\textbf{Simulation Comp.Time [ms]}} & \multicolumn{2}{c}{\textbf{Physical Comp. Time [ms]}} \\
    & $\mu_{c}$ & $\sigma_{c}$ & $\mu_{c}$ & $\sigma_{c}$ \\
    \midrule
    Frenet & 8.88 & 19.16 & 7.03 & 3.30 \\
    GBO & 16.03 & 4.58 & 9.20 & 4.87 \\
    Spliner & \textbf{2.31} & \textbf{2.86} & \textbf{3.65} & \textbf{1.49}\\
    Pred. Spliner & 40.17 & 33.81 & 45.70 & 39.80 \\
    FSDP \textbf{(ours)} & 10.43 & 31.33 & 9.84 & 22.46 \\
    \bottomrule
    \end{tabular}%
    }
    \caption{Computational results of each overtaking strategy. The computational times are reported with their mean $\mu_c$ and standard deviation $\sigma_c$.}
    \label{tab:compute}
    \vspace{-1.0em}
\end{table}

The computation time of \gls{FSDP} is higher than that of \emph{Spliner} and \emph{Frenet}, and is roughly comparable to that of \gls{gbo}. In terms of trajectory performance, \gls{FSDP} far exceeds the aforementioned three methods. \gls{FSDP} achieves higher performance with a slight increase in solution time, and the results are clearly evident. Compared to the high-performing \emph{Predictive Spliner}, \gls{FSDP}'s computation time was reduced by 74.04\% due to its use of \gls{sgp}. This demonstrates that \gls{FSDP} achieves an excellent balance between performance and computation time, exhibiting outstanding performance in overtaking scenarios.

\section{Conclusion} \label{sec:conclusion}
In this paper, we propose a data-driven overtaking algorithm for head-to-head autonomous racing competitions, named \gls{FSDP}. 
\gls{FSDP} first utilizes \gls{sgp} to predict opponent behavior, and then employs a bi-level \gls{qp} framework to generate the overtaking trajectory. 
Unlike previous approaches, \gls{FSDP} ensures the vehicle’s dynamics and safety while maintaining real-time performance by formulating the entire optimization problem in the Frenet frame within an \gls{mpc} framework.
Experiment results demonstrate that our method surpasses \gls{sota} by achieving an 8.93\% increase in overtaking success rate, supporting the maximum opponent speed among these methods, ensuring a smoother ego trajectory, and reducing 74.04\% computation time compared to the \emph{Predictive Spliner}, demonstrating its effectiveness in real-time motion planning.
Future work will focus on extending our approach to scenarios involving multiple opponents, where the complexity of the planning and prediction tasks increases. 


\bibliographystyle{IEEEtran}
\bibliography{reference}

\end{document}